%% file: main_icml22.tex
\documentclass[nohyperref]{article}

\usepackage{microtype}
\usepackage{graphicx}
\usepackage{subfigure}
\usepackage{booktabs} 

\usepackage{hyperref}

\usepackage[accepted]{icml2022}

\usepackage{mathtools}

\usepackage{cleveref}
\crefname{equation}{Eq.}{Eqs.}
\crefname{table}{Table}{Tables}
\crefname{figure}{Figure}{Figures}
\crefname{section}{Section}{Sections}
\crefname{algorithm}{Algorithm}{Algorithms}

\usepackage{url}
\usepackage{multirow,multicol,xspace}
\usepackage{amsthm,amsmath,amssymb}
\usepackage{float}
\usepackage{booktabs}
\usepackage{graphicx}
\usepackage{graphics}
\usepackage{paralist}
\usepackage{wrapfig}
\usepackage[normalem]{ulem}

\DeclareMathOperator*{\argmin}{arg\,min}

\newcommand*{\scale}[2][4]{\scalebox{#1}{$#2$}}

\theoremstyle{plain}

\theoremstyle{definition}

\theoremstyle{remark}

\usepackage[textsize=tiny]{todonotes}

\icmltitlerunning{Learning from Counterfactual Links for Link Prediction}

\begin{document}

\twocolumn[
\icmltitle{Learning from Counterfactual Links for Link Prediction}




\begin{icmlauthorlist}
\icmlauthor{Tong Zhao}{yyy}
\icmlauthor{Gang Liu}{yyy}
\icmlauthor{Daheng Wang}{yyy}
\icmlauthor{Wenhao Yu}{yyy}
\icmlauthor{Meng Jiang}{yyy}
\end{icmlauthorlist}

\icmlaffiliation{yyy}{Department of Computer Science and Engineering, University of Notre Dame, IN, USA}

\icmlcorrespondingauthor{Tong Zhao}{tzhao2@nd.edu}

\icmlkeywords{Machine Learning, ICML}

\vskip 0.3in
]



\printAffiliationsAndNotice{}  

\input{def}

\begin{abstract}
\input{body_icml/0abs}

\end{abstract}

\section{Introduction}
\label{sec:introduction}
\input{body_icml/1intro}

\section{Problem Definition}
\label{sec:preliminary}
\input{body_icml/2preliminary}

\section{Proposed Method}
\label{sec:method}

\input{body_icml/3method}

\section{Experiments}
\label{sec:experiments}

\input{body_icml/5experiment}

\section{Related Work}
\label{sec:related}
\input{body_icml/6related}

\section{Conclusion and Future Work}
\label{sec:conclusion}
\input{body_icml/7conclusion}

\section*{Acknowledgements}
This research was supported by NSF Grants IIS-1849816, IIS-2142827, and IIS-2146761.


\bibliography{ref}
\bibliographystyle{icml2022}

\newpage
\appendix
\onecolumn
\input{body_icml/8appendix}

\end{document}

%% file: def.tex
\newcommand{\method}{\textsc{CFLP}\xspace}

\newcommand{\cora}{\textsc{Cora}\xspace}
\newcommand{\cseer}{\textsc{CiteSeer}\xspace}
\newcommand{\pubmed}{\textsc{PubMed}\xspace}
\newcommand{\ddi}{\textsc{OGB-ddi}\xspace}
\newcommand{\facebook}{\textsc{Facebook}\xspace}

\newcommand\tong[1]{\textcolor{cyan}{[Tong: #1]}}
\newcommand\gang[1]{\textcolor{green}{[Gang: #1]}}
\newcommand{\wyu}[1]{\textcolor{magenta}{wyu: #1}}

%% file: body_icml/0abs.tex
Learning to predict missing links is important for many graph-based applications. Existing methods were designed to learn the association between observed graph structure and existence of link between a pair of nodes. However, the causal relationship between the two variables was largely ignored for learning to predict links on a graph. In this work, we visit this factor by asking a counterfactual question: ``\emph{would the link still exist if the graph structure became different from observation?}'' Its answer, counterfactual links, will be able to augment the graph data for representation learning. To create these links, we employ causal models that consider the information (i.e., learned representations) of node pairs as context, global graph structural properties as treatment, and link existence as outcome. We propose a novel data augmentation-based link prediction method that creates counterfactual links and learns representations from both the observed and counterfactual links. Experiments on benchmark data show that our graph learning method achieves state-of-the-art performance on the task of link prediction.

%% file: body_icml/1intro.tex
Link prediction seeks to predict the likelihood of edge existence between node pairs based on observed graph. Given the omnipresence of graph-structured data, link prediction has copious applications, such as movie recommendation~\citep{bennett2007netflix}, chemical interaction prediction~\citep{stanfield2017drug}, and knowledge graph completion~\citep{kazemi2018simple}. Graph machine learning methods have been widely applied to solve this problem. Their standard scheme is to first learn representation vectors of nodes and then learn the \emph{association} between the representations of a pair of nodes and the existence of link between them. For example, graph neural networks (GNNs) use neighborhood aggregation to create the representation vectors: the representation vector of a node is computed by recursively aggregating and transforming representation vectors of its neighboring nodes~\citep{kipf2016semi,hamilton2017inductive,wu2020comprehensive}. Then the vectors are fed into a binary classification model to learn the \emph{association}. GNN methods have shown predominance in the task of link prediction~\citep{zhang2020revisiting}.

\begin{figure*}[t]
    \centering
    \subfigure[Find counterfactual link as the most similar node pair with a different treatment.]
    {\includegraphics[width=0.4\linewidth]{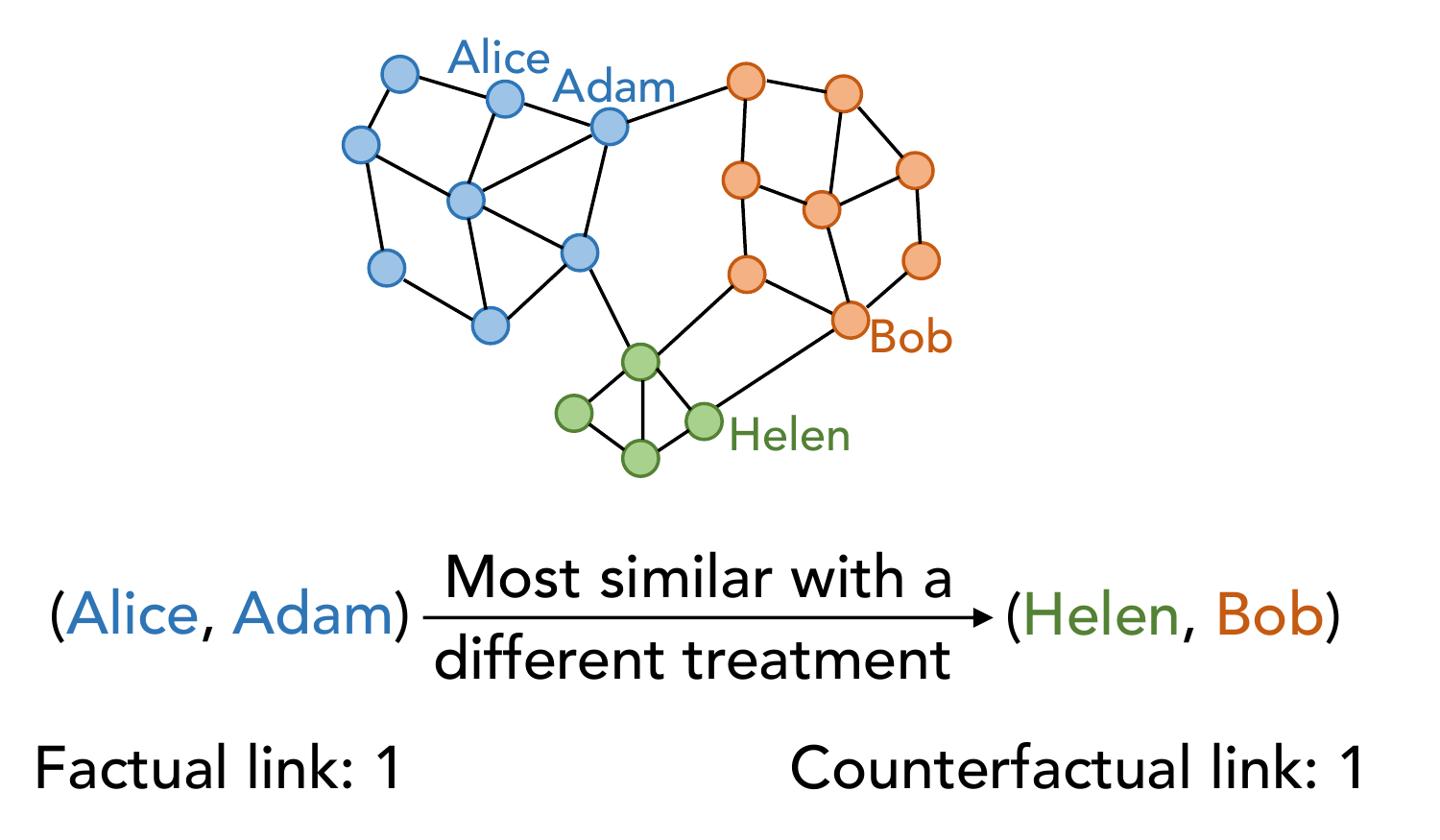}\label{fig:framework1}}
    \hfill
    \subfigure[Train a GNN-based link predictor to predict factual and counterfactual links given the corresponding treatments.]
    {\includegraphics[width=0.57\linewidth]{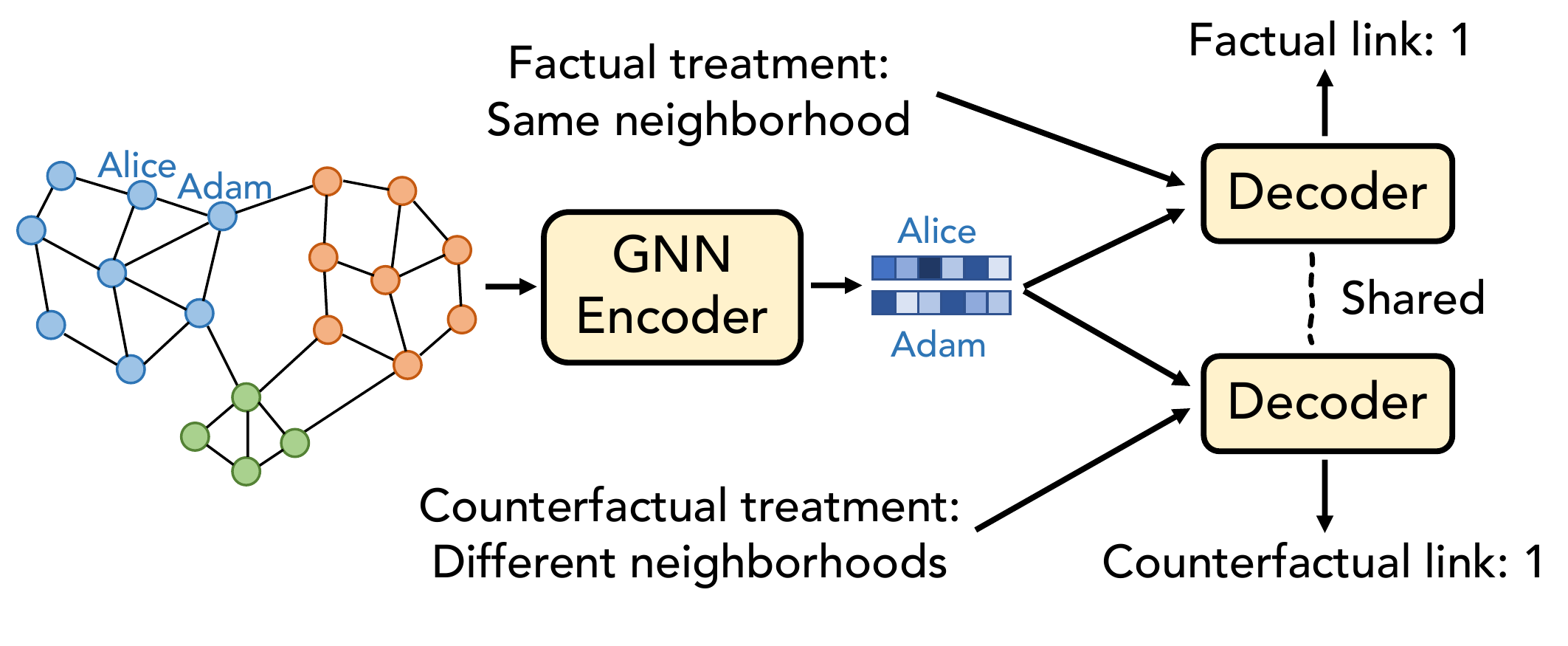}\label{fig:framework2}}
    \caption{The proposed \method learns the causal relationship between the observed graph structure (e.g., neighborhood similarity, considered as treatment variable) and link existence (considered as outcome). In this example, the link predictor would be trained to estimate the individual treatment effect (ITE) as $1-1=0$ so it looks for factors other than neighborhood to predict the factual link.}
    \label{fig:framework}
\end{figure*}

Unfortunately, the causal relationship between graph structure and link existence was largely ignored in previous work. Existing methods that learn from association are not able to capture essential factors to accurately predict missing links in \emph{test data}. Take a specific social network as an example. Suppose Alice and Adam live in the same neighborhood and they are close friends. The association between neighborhood belonging and friendship could be too strong to discover the essential factors of friendship such as common interests or family relationships. Such factors could also be the cause of them living in the same neighborhood. So, our idea is to ask a \emph{counterfactual} question: ``\emph{would Alice and Adam still be close friends if they were not living in the same neighborhood?}'' If a graph learning model can learn the causal relationship by answering this counterfactual question, it will improve the accuracy of link prediction with such knowledge. Generally, the questions can be described as ``\emph{would the link exist or not if the graph structure became different from observation?}''

As known to many, counterfactual questions are the key component of causal inference and have been well defined in literature. A counterfactual question is usually framed with three factors: context (as a data point), manipulation (e.g., treatment, intervention, action, strategy), and outcome \citep{van2007causal,johansson2016learning}. (To simplify the language, we use ``treatment'' to refer to the manipulation in this paper, as readers might be familiar more with the word ``treatment.'') Given certain data context, it asks what the outcome would have been if the treatment had not been the observed value. In the scenario of link prediction, we consider the information of a pair of nodes as context, graph structural properties as treatment, and link existence as outcome. Recall the social network example. The context is the representations of Alice and Adam that are learned from their personal attributes and relationships with others on the social network. The treatment is whether live in the same neighborhood, which can be identified by community detection. And the outcome is their friendship.

In this work, we propose a novel concept of ``counterfactual link'' that answers the counterfactual question and (based on this concept) a novel link prediction method (\method) that uses the counterfactual links as augmented data for graph representation learning.
Figure~\ref{fig:framework} illustrates this two-step method.
Suppose the treatment variable is defined as one type of global graph structure, e.g., the neighborhood assignment discovered by spectral clustering or community detection algorithms.
We are wondering how likely the neighborhood distribution makes a difference on the link (non-)existence for each pair of nodes.
So, given a pair of nodes (like Alice and Adam) and the treatment value on this pair (in the same neighborhood), we find a pair of nodes (like Helen and Bob) that satisfies two conditions: (1) it has a different treatment (in different neighborhoods) and (2) it is the most similar pair with the given pair of nodes. 
We name these matched pairs of nodes as counterfactual links. Note that the outcome of the counterfactual links can be either 1 or 0, depending on whether there exists an edge between the matched pair of nodes.
The counterfactual link provides an unobservable outcome to the given pair of nodes under a counterfactual condition.
The process of creating counterfactual links for all positive and negative training examples can be viewed as a graph data augmentation method, as it enriches the training set.
Then, \method trains a link predictor (which is GNN-based) to learn the representation vectors of nodes to predict both the observed factual links and counterfactual links.
In this Alice-Adam example, the link predictor is trained to estimate the individual treatment effect (ITE) of neighborhood assignment as $1-1=0$, where ITE is a metric for the effect of treatment on the outcome and zero indicates the given treatment has no effect on the outcome. 
So, the learner will try to discover the essential factors on the friendship between Alice and Adam.
\method learns from the counterfactual links to find these factors for graph learning models to accurately predict missing links.

\textbf{Contributions.} Our main contributions can be summarized as follows.
(1) This is the first work that aims at improving link prediction by causal inference, specifically, generating counterfactual links to answer counterfactual questions about link existence.
(2) This work introduces \method that trains GNN-based link predictors to predict both factual and counterfactual links. 
It leverages causal relationship between global graph structure and link existence to enhance link prediction.
(3) \method outperforms competitive baselines on several benchmark datasets. 
We analyze the impact of counterfactual links as well as the choice of treatment variable. This work sheds insights for improving graph machine learning with causal analysis, which has not been extensively studied yet, while the other direction (machine learning for causal inference) has been studied for long.
Source code of the proposed CFLP method is publicly available at \url{https://github.com/DM2-ND/CFLP}.

%% file: body_icml/2preliminary.tex
\paragraph{Notations}
Let $G = (\mathcal{V}, \mathcal{E})$ be an undirected graph of $N$ nodes, where $\mathcal{V} = \{v_1, v_2, \dots, v_N\}$ is the set of nodes and $\mathcal{E} \subseteq \mathcal{V} \times \mathcal{V}$ is the set of observed links. We denote the adjacency matrix as $\mathbf{A} \in \{0, 1\}^{N \times N}$, where $A_{i,j} = 1$ indicates nodes $v_i$ and $v_j$ are connected and vice versa. We denote the node feature matrix as $\mathbf{X} \in \mathbb{R}^{N \times F}$, where $F$ is the number of node features and $\mathbf{x}_{i}$ indicates the feature vector of node $v_i$ (the $i$-th row of $\mathbf{X}$). 

In this work, we follow the commonly accepted problem definition of link prediction on graph data~\citep{zhang2018link,zhang2020revisiting,cai2021line}: Given an observed graph $G$ (with validation and testing links masked off), predict the link existence between every pair of nodes. More specifically, for the GNN-based link prediction methods, they learn low-dimensional node representations $\mathbf{Z} \in \mathbb{R}^{N \times H}$, where $H$ is the dimensional size of latent space such that $H \ll N$, and then use $\mathbf{Z}$ for the prediction of link existence between every node pair.

%% file: body_icml/3method.tex
\subsection{Improving Graph Learning with Causal Model}
\label{sec:setup}

\begin{figure}
    \centering
    \subfigure[Causal modeling (not the target of our work but related to the idea we propose): Given $\mathbf{Z}$ and observed outcomes, find treatment effect of $T$ on $Y$.]
    {\includegraphics[width=0.47\linewidth]{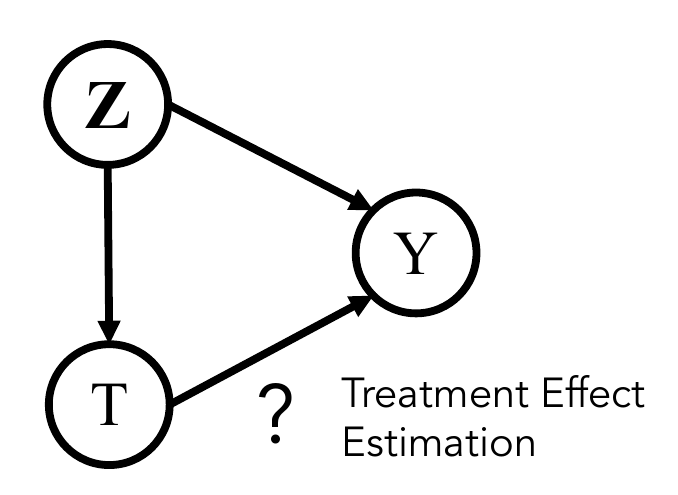}\label{fig:dag1}}
    \hfill
    \subfigure[Graph learning with causal model (the proposed idea): leverage the estimated $\text{ITE}(A_{i,j}|T_{i,j})$ to improve the learning of $\mathbf{z}_i$ and $\mathbf{z}_j$.]
    {\includegraphics[width=0.47\linewidth]{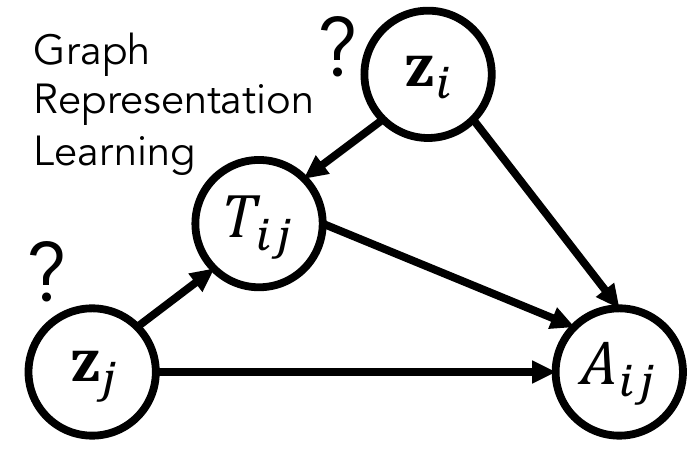}\label{fig:dag2}}
    \vspace{-0.05in}
    \caption{Our proposed work improves graph representation learning by leveraging causal model.}
    \vspace{-0.1in}
\end{figure}

\paragraph{Leveraging Causal Model(s)} Counterfactual causal inference aims to find out the causal relationship between treatment and outcome by asking the counterfactual questions such as ``would the outcome be different if the treatment was different?''~\citep{morgan2015counterfactuals}. \cref{fig:dag1} is a typical example, in which we denote the context (confounder) as $\mathbf{Z}$, treatment as $T$, and the outcome as $Y$. Given the context, treatments, and their corresponding outcomes, counterfactual inference methods aim to find the effect of treatment on the outcome, which is usually measured by \emph{individual treatment effect} (ITE) and its expectation \emph{averaged treatment effect} (ATE)~\citep{van2007causal,weiss2015machine}. For a binary treatment variable $T = \{0, 1\}$, denoting $g(\mathbf{z}, T)$ as the outcome of $\mathbf{z}$ given the treatment $T$, we have $\text{ITE}(\mathbf{z}) = g(\mathbf{z}, 1) - g(\mathbf{z}, 0)$, and $\text{ATE} = \mathbb{E}_{\mathbf{z}\sim \mathbf{Z}}~\text{ITE}(\mathbf{z})$.

Ideally, we need all potential outcomes of the contexts under all kinds of treatments to study the causal relationships~\citep{morgan2015counterfactuals}. However, in reality, the fact that we can only observe the outcome under one particular treatment prevents the ITE from being known~\citep{johansson2016learning}. Traditional causal inference methods use statistical learning approaches such as Neyman–Rubin 
causal model (BCM) and propensity score matching (PSM) to predict the value of ATE~\citep{rubin1974estimating,rubin2005causal}.

In this work, we look at \emph{link prediction} with graph learning, which is to learn effective node representations $\mathbf{Z}$ for predicting link existence in test data. In \cref{fig:dag2}, $\mathbf{z}_i$ and $\mathbf{z}_j$ are representations of nodes $v_i$ and $v_j$, and the outcome $A_{i,j}$ is the link existence between $v_i$ and $v_j$.
Here, the objective is different from classic causal inference. In graph learning, we want to improve the learning of $\mathbf{z}_i$ and $\mathbf{z}_j$ with the estimation on the effect of treatment $T_{i,j}$ on the outcome $A_{i,j}$. Specifically, for each pair of nodes $(v_i, v_j)$, its ITE can be estimated by
\vspace{-0.05in}
\begin{equation}
\label{eq:ite}
    \text{ITE}_{(v_i,v_j)} = g((\mathbf{z}_i, \mathbf{z}_j), 1) - g((\mathbf{z}_i, \mathbf{z}_j), 0)
\vspace{-0.03in}
\end{equation}
and we use this information to improve the learning of $\mathbf{Z}$.

We denote by $\mathbf{A}$ the observed adjacency matrix as the \emph{factual} outcomes, and denote by $\mathbf{A}^{CF}$ the unobserved matrix of the counterfactual links when the treatment is different as the \emph{counterfactual} outcomes. We denote $\mathbf{T} \in \{0, 1\}^{N \times N}$ as the binary factual treatment matrix, where $T_{i, j}$ indicates the treatment of the node pair $(v_i, v_j)$. We denote $\mathbf{T}^{CF}$ as the counterfactual treatment matrix where $T^{CF}_{i, j} = 1 - T_{i, j}$. We are interested in (a) estimating the counterfactual outcomes $\mathbf{A}^{CF}$ and (b) learning from both factual and counterfactual outcomes $\mathbf{A}$ and $\mathbf{A}^{CF}$ (as observed and augmented data) to enhance link prediction.

\vspace{-0.05in}
\paragraph{Treatment Variable}
Previous works on GNN-based link prediction~\citep{zhang2018link,zhang2020revisiting} have shown that the message passing-based GNNs are capable to capture the structural information (e.g., Katz index) for link prediction. Nevertheless, as illustrated by the Alice-Adam example in \cref{sec:introduction}, the association between such structural information and actual link existence may be too strong for models to discover more essential factors than it, hence resulting in sub-optimal link prediction performance. 
Therefore, in this work, we use the global structural role of each node pair as its treatment. It's worth mentioning that the causal model shown in \cref{fig:dag2} does not limit the treatment to be structural roles, i.e., $T_{i, j}$ can be any binary property of node pair $(v_i, v_j)$.
Without the loss of generality, we use Louvain~\citep{blondel2008fast}, an unsupervised approach that has been widely used for community detection, as an example. Louvain discovers community structure of a graph and assigns each node to one community. Then we can define the binary treatment variable as whether these two nodes in the pair belong to the same community.
Let $c: \mathcal{V} \rightarrow \mathbb{N}$ be any graph mining/clustering method that outputs the index of community/cluster/neighborhood that each node belongs to. The treatment matrix $\mathbf{T}$ is defined as $T_{i,j} = 1$ if $c(v_i) = c(v_j)$, and $T_{i,j} = 0$ otherwise.
For the choice of $c$, we suggest methods that group nodes based on global graph structural information, including but not limited to Louvain~\citep{blondel2008fast}, K-core~\citep{bader2003automated}, and spectral clustering~\citep{ng2001spectral}.

\subsection{Counterfactual Links} 
\label{sec:cflinks}
\vspace{-0.05in}
To implement the solution based on above idea, we propose counterfactual links.
As aforementioned, for each node pair, the observed data contains only the factual treatment and outcome, meaning that the link existence for the given node pair with an opposite treatment is unknown. Therefore, we use the outcome from the nearest observed context as a substitute. 
This type of matching on covariates is widely used to estimate treatment effects from observational data~\citep{johansson2016learning,alaa2019validating}.
That is, we want to find the nearest neighbor with the opposite treatment for each observed node pairs and use the nearest neighbor's outcome as a \emph{counterfactual link}. 
Formally, $\forall (v_i, v_j) \in \mathcal{V} \times \mathcal{V}$,  its counterfactual link $(v_a, v_b)$ is
\vspace{-0.05in}
\begin{equation}
\label{eq:nearest1}
    (v_a, v_b) = \argmin_{v_a, v_b \in \mathcal{V}} \{h((v_i, v_j), (v_a, v_b)) \text{ } | \text{ } T_{a, b} = 1- T_{i, j} \},
\end{equation}
where $h(\cdot, \cdot)$ is a metric of measuring the distance between a pair of node pairs (a pair of contexts). Nevertheless, finding the nearest neighbors by computing the distance between all pairs of node pairs is extremely inefficient and infeasible in application, which takes $O(N^4)$ comparisons (as there are totally $O(N^2)$ node pairs). 
Hence we implement \cref{eq:nearest1} using node-level embeddings. 
Specifically, considering that we want to find the nearest node pair based on both the raw node features and structural features, 
we take the state-of-the-art unsupervised graph representation learning method MVGRL~\citep{hassani2020contrastive} to learn the node embeddings $\tilde{\mathbf{X}} \in \mathbb{R}^{N \times \tilde{F}}$ from the observed graph (with validation and testing links masked off). We use $\tilde{\mathbf{X}}$ to find the nearest neighbors of node pairs. Therefore, $\forall (v_i, v_j) \in \mathcal{V} \times \mathcal{V}$, we define its counterfactual link $(v_a, v_b)$ as
\vspace{-0.05in}
\begin{align}
    \label{eq:nearest2}
    (v_a, v_b) = &\argmin_{v_a, v_b \in \mathcal{V}} \{d(\tilde{\mathbf{x}}_{i}, \tilde{\mathbf{x}}_{a})+d(\tilde{\mathbf{x}}_{j}, \tilde{\mathbf{x}}_{b}) \text{ } | \text{ }  \\
    &T_{a, b} = 1- T_{i, j}, d(\tilde{\mathbf{x}}_{i}, \tilde{\mathbf{x}}_{a})+d(\tilde{\mathbf{x}}_{j}, \tilde{\mathbf{x}}_{b}) < 2\gamma \}, \nonumber
\vspace{-0.02in}
\end{align}
where $d(\cdot, \cdot)$ is specified as the Euclidean distance on the embedding space of $\tilde{\mathbf{X}}$, and $\gamma$ is a hyperparameter that defines the maximum distance that two nodes are considered as similar. When no node pair satisfies the above equation (i.e., there does not exist any node pair with opposite treatment that is close enough to the target node pair), we do not assign any nearest neighbor for the given node pair to ensure all the neighbors are similar enough (as substitutes) in the feature space. Thus, the counterfactual treatment matrix $\mathbf{T}^{CF}$ and the counterfactual adjacency matrix $\mathbf{A}^{CF}$ are defined as
\begin{equation}
\label{eq:tcf}
    T^{CF}_{i, j}, A^{CF}_{i, j} = 
    \begin{cases}
        1 - T_{i, j}, A_{a, b} & \text{, if } \exists$ $(v_a, v_b) \in \mathcal{V} \times \mathcal{V} \\
         & \quad \text{ satisfies \cref{eq:nearest2};} \\
        T_{i,j}, A_{i,j} &  \text{, otherwise}. 
    \end{cases}
\end{equation}
It is worth noting that the node embeddings $\tilde{\mathbf{X}}$ and the nearest neighbors are computed only once and do not change during the learning process. 
$\tilde{\mathbf{X}}$ is only used for finding the nearest neighbors.

\paragraph{Learning from Counterfactual Distributions}
Let $P^{F}$ be the {factual distribution} of the observed contexts and treatments, and $P^{CF}$ be the {counterfactual distribution} that is composed of the observed contexts and {opposite} treatments. We define the {empirical} factual distribution $\hat{P}^{F} \sim P^{F}$ as $\hat{P}^{F} = \{(v_i, v_j, T_{i,j})\}^N_{i,j=1}$, and define the empirical counterfactual distribution $\hat{P}^{CF} \sim P^{CF}$ as $\hat{P}^{CF} = \{(v_i, v_j, T^{CF}_{i,j})\}^N_{i,j=1}$. 
Unlike traditional link prediction methods that take only $\hat{P}^{F}$ as input and use the observed outcomes $\mathbf{A}$ as the training target, 
we take advantage of the counterfactual distribution by using it as the augmented training data.
That is, we use $\hat{P}^{CF}$ as a complementary input and use the counterfactual outcomes $\mathbf{A}^{CF}$ as the training target for the counterfactual data samples.

\subsection{Learning from Counterfactual Links} 
\label{sec:model}

In this subsection, we present the design of our model as well as the training method.
The input of the model in \method includes (1) the observed graph data $\mathbf{A}$ and raw feature matrix $\mathbf{X}$, (2) the factual treatments $\mathbf{T}^F$ and counterfactual treatments $\mathbf{T}^{CF}$, and (3) the counterfactual links data $\mathbf{A}^{CF}$.
The output contains link prediction logits in $\widehat{\mathbf{A}}$ and $\widehat{\mathbf{A}}^{CF}$ for the factual and counterfactual adjacency matrices $\mathbf{A}$ and $\mathbf{A}^{CF}$, respectively.

\paragraph{Graph Learning Model}
The model consist of two trainable components: a graph encoder $f$ and a link decoder $g$.
The graph encoder generates representation vectors of nodes from graph data.
And the link decoder projects the representation vectors of node pairs into the link prediction logits.
The choice of the graph encoder $f$ can be any end-to-end GNN model. Without the loss of generality, here we use the commonly used graph convolutional network (GCN)~\citep{kipf2016semi}. Each layer of GCN is defined as
\vspace{-0.05in}
\begin{equation}
\scale[0.9]{
    \mathbf{H}^{(l)} = f^{(l)}(\mathbf{A}, \mathbf{H}^{(l-1)}; \mathbf{W}^{(l)}) = \sigma(\tilde{\mathbf{D}}^{-\frac{1}{2}}\tilde{\mathbf{A}}\tilde{\mathbf{D}}^{-\frac{1}{2}}\mathbf{H}^{(l-1)}\mathbf{W}^{(l)}),
    }
    \label{eq:gcn_layer}
\vspace{-0.02in}
\end{equation}
where $l$ is the layer index, $\tilde{\mathbf{A}} = \mathbf{A} + \mathbf{I}$ is the adjacency matrix with added self-loops,  $\tilde{\mathbf{D}}$ is the diagonal degree matrix $\tilde{D}_{ii} = \sum_j \tilde{A}_{ij}$, $\mathbf{H}^{(0)} = \mathbf{X}$, $\mathbf{W}^{(l)}$ is the learnable weight matrix at the $l$-th layer, and $\sigma(\cdot)$ denotes the nonlinear activation ReLU. We denote $\mathbf{Z} = f(\mathbf{A}, \mathbf{X}) \in \mathbb{R}^{N \times H}$ as the output from the encoder's last layer, i.e., the $H$-dimensional representation vectors of nodes. Following previous works~\citep{zhang2018link,zhang2020revisiting}, we compute the representation of a node pair as the Hadamard product of the vectors of the two nodes. That is, the representation for the node pair $(v_i, v_j)$ is $\mathbf{z}_i \odot \mathbf{z}_j \in \mathbb{R}^{H}$, where $\odot$ stands for the Hadamard product.

For the link decoder that predicts whether a link exists between a pair of nodes, we opt for simplicity and adopt a simple decoder based on multi-layer perceptron (MLP), given the representations of node pairs and their treatments. That is, the decoder $g$ is defined as
\begin{align}
    \label{eq:apredf}
    &\scale[0.94]{\widehat{\mathbf{A}} = g(\mathbf{Z}, \mathbf{T}), \text{ s.t. } \widehat{A}_{i, j} = \text{MLP}([\mathbf{z}_i \odot \mathbf{z}_j, T_{i, j}]),} \\
    \label{eq:apredcf}
    &\scale[0.94]{\widehat{\mathbf{A}}^{CF} = g(\mathbf{Z}, \mathbf{T}^{CF}), \text{ s.t. } \widehat{A}^{CF}_{i, j} = \text{MLP}([\mathbf{z}_i \odot \mathbf{z}_j, T^{CF}_{i, j}]),}
\end{align}
where $[\cdot, \cdot]$ stands for the concatenation of vectors, and $\widehat{\mathbf{A}}$ and $\widehat{\mathbf{A}}^{CF}$ can also be used for estimating the observed ITE as aforementioned in \cref{eq:ite}.

During the training process, data samples from the empirical factual distribution $\hat{P}^{F}$ and the empirical counterfactual distribution $\hat{P}^{CF}$ are fed into decoder $g$ and optimized towards $\mathbf{A}$ and $\mathbf{A}^{CF}$, respectively.
That is, for the two distributions, the loss functions are as follows:
\begin{align}
    \label{eq:eplossf}
    \mathcal{L}_{F} = &\frac{1}{N^2}\sum_{i=1}^N\sum_{j=1}^N A_{i,j} \cdot \log \widehat{A}_{i,j} \\
    &+ (1 - A_{i,j}) \cdot \log(1 - \widehat{A}_{i,j}), \nonumber\\
    \label{eq:eplosscf}
    \mathcal{L}_{CF} = &\frac{1}{N^2}\sum_{i=1}^N\sum_{j=1}^N A^{CF}_{i,j} \cdot \log\widehat{A}^{CF}_{i,j} \\
    &+ (1 - A^{CF}_{i,j}) \cdot \log(1 - \widehat{A}^{CF}_{i,j}). \nonumber
\end{align}

\paragraph{Balancing Counterfactual Learning}
In the training process, the above loss minimizations train the model on both the empirical factual distribution $\hat{P}^{F} \sim P^{F}$ and empirical counterfactual distribution $\hat{P}^{CF} \sim P^{CF}$ that are not necessarily equal -- the training examples (node pairs) do not have to be aligned.
However, at the stage of inference, the test data contains only observed (factual) samples. Such a gap between the training and testing data distributions exposes the model in the risk of covariant shift, which is a common issue in counterfactual representation learning~\citep{johansson2016learning,assaad2021counterfactual}.

To force the distributions of representations of factual distributions and counterfactual distributions to be similar, we adopt the discrepancy distance~\citep{mansour2009domain,johansson2016learning} as another training objective to regularize the representation learning. That is, we use the following loss term to minimize the distance between the learned representations from $\hat{P}^{F}$ and $\hat{P}^{CF}$:
\begin{equation}
    \label{eq:discloss}
    \mathcal{L}_{disc} = \text{disc}(\hat{P}^{F}_f, \hat{P}^{CF}_f), \text{ where } \text{disc}(P, Q) = ||P - Q||_{F},
\end{equation}
where $||\cdot||_F$ denotes the Frobenius Norm, and $\hat{P}^{F}_f$ and $\hat{P}^{CF}_f$ denote the node pair representations learned by graph encoder $f$ from factual distribution and counterfactual distribution, respectively. Specifically, the learned representations for $(v_i, v_j, T_{i,j})$ and $(v_i, v_j, T_{i,j}^{CF})$ are $[\mathbf{z}_i \odot \mathbf{z}_j, T_{i,j}]$ (\cref{eq:apredf}) and $[\mathbf{z}_i \odot \mathbf{z}_j, T_{i,j}^{CF}]$ (\cref{eq:apredcf}), respectively.

\paragraph{Training}
During the training of \method, we want the model to be optimized towards three targets: (1) accurate link prediction on the observed outcomes (\cref{eq:eplossf}), (2) accurate prediction on the counterfactual links (\cref{eq:eplosscf}), and (3) regularization on the representation spaces learned from $\hat{P}^{F}$ and $\hat{P}^{CF}$ (\cref{eq:discloss}). Therefore, the overall training loss of our proposed \method is
\begin{equation}
    \mathcal{L} = \mathcal{L}_{F} + \alpha \cdot \mathcal{L}_{CF} + \beta \cdot \mathcal{L}_{disc},
    \label{eq:loss}
\end{equation}
where $\alpha$ and $\beta$ are hyperparameters to control the weights of counterfactual outcome estimation (link prediction) loss and discrepancy loss.

\begin{algorithm}[tb]
   \caption{\method}
   \label{alg:method}
\begin{algorithmic}
    \STATE {\bfseries Input:} $f$, $g$, $\mathbf{A}$, $\mathbf{X}$, $n\_epochs$, $n\_epoch\_ft$
    \STATE Compute $\mathbf{T}$ as presented in \cref{sec:setup}.
    \STATE Compute $\mathbf{T}^{CF}, \mathbf{A}^{CF}$ by \cref{eq:nearest2,eq:tcf}.
    \STATE \verb+// model training+
    \FOR{epoch in range($n\_epochs$)}
        \STATE $\mathbf{Z} = f(\mathbf{A}, \mathbf{X})$.
        \STATE Get $\widehat{\mathbf{A}}$ and $\widehat{\mathbf{A}}^{CF}$ via $g$ with \cref{eq:apredf,eq:apredcf}.
        \STATE Update $\Theta_f$ and $\Theta_g$ with $\mathcal{L}$. (\cref{eq:loss})
    \ENDFOR
    \STATE \verb+// decoder fine-tuning+
    \STATE Freeze $\Theta_f$ and re-initialize $\Theta_g$.
    \STATE $\mathbf{Z} = f(\mathbf{A}, \mathbf{X})$.
    \FOR{epoch in range($n\_epochs\_ft$)}
        \STATE Get $\widehat{\mathbf{A}}$ via $g$ with \cref{eq:apredf}.
        \STATE Update $\Theta_g$ with $\mathcal{L}_F$. (\cref{eq:eplossf})
    \ENDFOR
    \STATE \verb+// inference+
    \STATE $\mathbf{Z} = f(\mathbf{A}, \mathbf{X})$.
    \STATE Get $\widehat{\mathbf{A}}$ and $\widehat{\mathbf{A}}^{CF}$ via $g$ with \cref{eq:apredf,eq:apredcf}.
    \STATE {\bfseries Output:} $\widehat{\mathbf{A}}$ for link prediction, $\widehat{\mathbf{A}}^{CF}$.
\end{algorithmic}
\end{algorithm}

\paragraph{Summary}
\cref{alg:method} summarizes the whole process of \method. The \textbf{first step} is to compute the factual and counterfactual treatments $\mathbf{T}$, $\mathbf{T}^{CF}$ as well as the counterfactual links $\mathbf{A}^{CF}$. Then, the \textbf{second step} trains the graph learning model on both the observed factual link existence and generated counterfactual link existence with the integrated loss function (\cref{eq:loss}). Note that the discrepancy loss (\cref{eq:discloss}) is computed on the representations of node pairs learned by the graph encoder $f$, so the decoder $g$ is trained with data from both $\hat{P}^{F}$ and $\hat{P}^{CF}$ without balancing the constraints. Therefore, after the model is sufficiently trained, we freeze the graph encoder $f$
and fine-tune $g$ with only the factual data. Finally, after the decoder is sufficiently fine-tuned, we output the link prediction logits for both the factual and counterfactual adjacency matrices.

\begin{table*}[t]
\small
  \caption{Statistics of datasets used in the experiments.}
  \label{tab:data}
  \centering
  \begin{tabular}{lrrrrr}
    \toprule
    Dataset & \cora & \cseer & \pubmed & \facebook & \ddi \\
    \midrule
    \# nodes & 2,708 & 3,327 & 19,717 & 4,039 & 4,267  \\
    \# links & 5,278 & 4,552 & 44,324 & 88,234 & 1,334,889	\\
    \# validation node pairs & 1,054 & 910 & 8,864 & 17,646 & 235,371 \\
    \# test node pairs & 2,110 & 1,820 & 17,728 & 35,292 & 229,088 \\
    \bottomrule
  \end{tabular}
\end{table*}

\paragraph{Complexity}
The complexity of the first step (finding counterfactual links with nearest neighbors) is proportional to the number of node pairs. When $\gamma$ is set as a small value to obtain indeed similar node pairs, this step (\cref{eq:nearest2}) uses constant time. Moreover, the computation in \cref{eq:nearest2} can be parallelized. Therefore, the time complexity is $O(N^2/C)$ where $C$ is the number of processes.
For the complexity of the second step (training counterfactual learning model), the GNN encoder has time complexity of $O(L H^2 N + L H |\mathcal{E}|)$~\citep{wu2020comprehensive}, where $L$ is the number of GNN layers and $H$ is the size of node representations. Given that we sample the same number of non-existing links as that of observed links during training, the complexity of a \emph{three-layer MLP} decoder is $O(((H+1)\cdot d_h + d_h \cdot 1 ) |\mathcal{E}|) = O(d_h (H+2) |\mathcal{E}|)$, where $d_h$ is the number of neurons in the hidden layer.
Therefore, the second step has linear time complexity w.r.t. the sum of node and edge counts.

\paragraph{Limitations}
First, as mentioned above, the computation of finding counterfactual links has a worst-case complexity of $O(N^2)$. Second, \method performs counterfactual prediction with only a single treatment; however, there are quite a few kinds of graph structural information that can be considered as treatments. Future work can leverage the rich structural information by bundled treatments~\citep{zou2020counterfactual} in the generation of counterfactual links.

%% file: body_icml/5experiment.tex
\begin{table*}[t]
\small
  \caption{Link prediction performances measured by Hits@20. Best performance and best baseline performance are marked with bold and underline, respectively.}
  \label{tab:result_hits}
  \centering
  \begin{tabular}{lrrrrr}
    \toprule
    & \multicolumn{1}{c}{\cora} & \multicolumn{1}{c}{\cseer} & \multicolumn{1}{c}{\pubmed} & \multicolumn{1}{c}{\facebook} & \multicolumn{1}{c}{\ddi} \\
    \midrule
    Node2Vec & 49.96$\pm$2.51 & 47.78$\pm$1.72 & 39.19$\pm$1.02 & 24.24$\pm$3.02 & 23.26$\pm$2.09 \\
    MVGRL & 19.53$\pm$2.64 & 14.07$\pm$0.79 & 14.19$\pm$0.85 & 14.43$\pm$0.33 & 10.02$\pm$1.01 \\
    VGAE & 45.91$\pm$3.38 & 44.04$\pm$4.86 & 23.73$\pm$1.61 & 37.01$\pm$0.63 & 11.71$\pm$1.96 \\
    SEAL & 51.35$\pm$2.26 & 40.90$\pm$3.68 & 28.45$\pm$3.81 & 40.89$\pm$5.70 & 30.56$\pm$3.86 \\
    LGLP & \underline{62.98}$\pm$0.56 & \underline{57.43}$\pm$3.71 & \multicolumn{1}{c}{--} & 37.86$\pm$2.13 & \multicolumn{1}{c}{--}\\
    GCN & 49.06$\pm$1.72 & 55.56$\pm$1.32 & 21.84$\pm$3.87 & \underline{53.89}$\pm$2.14 & 37.07$\pm$5.07 \\
    GSAGE & 53.54$\pm$2.96 & 53.67$\pm$2.94 & \underline{39.13}$\pm$4.41 & 45.51$\pm$3.22 & 53.90$\pm$4.74 \\
    JKNet & 48.21$\pm$3.86 & 55.60$\pm$2.17 & 25.64$\pm$4.11 & 52.25$\pm$1.48  & \underline{60.56}$\pm$8.69\\
    \midrule
    \multicolumn{6}{l}{Our proposed \method with different graph encoders} \\
    \method w/ GCN & 60.34$\pm$2.33 & 59.45$\pm$2.30 & 34.12$\pm$2.72 & 53.95$\pm$2.29 & 52.51$\pm$1.09  \\
    \method w/ GSAGE & 57.33$\pm$1.73 & 53.05$\pm$2.07 & 43.07$\pm$2.36 & 47.28$\pm$3.00 & 75.49$\pm$4.33 \\
    \method w/ JKNet & \textbf{65.57}$\pm$1.05 & \textbf{68.09}$\pm$1.49 & \textbf{44.90}$\pm$2.00 & \textbf{55.22}$\pm$1.29 & \textbf{86.08}$\pm$1.98 \\
    \bottomrule
  \end{tabular}
  \vspace{-0.05in}
\end{table*}

\begin{table*}[t]
\small
  \caption{Link prediction performances measured by AUC. Best performance and best baseline performance are marked with bold and underline, respectively.}
  \label{tab:result_auc}
  \centering
  \begin{tabular}{lccccc}
    \toprule
    & \cora & \cseer & \pubmed & \facebook & \ddi \\
    \midrule
    Node2Vec & 84.49$\pm$0.49 & 80.00$\pm$0.68 & 80.32$\pm$0.29 & 86.49$\pm$4.32 & 90.83$\pm$0.02 \\
    MVGRL & 75.07$\pm$3.63 & 61.20$\pm$0.55 & 80.78$\pm$1.28 & 79.83$\pm$0.30 & 81.45$\pm$0.99 \\
    VGAE & 88.68$\pm$0.40 & 85.35$\pm$0.60 & 95.80$\pm$0.13 & 98.66$\pm$0.04 & 93.08$\pm$0.15 \\
    SEAL & \underline{92.55}$\pm$0.50 & 85.82$\pm$0.44 & 96.36$\pm$0.28 & \textbf{\underline{99.60}}$\pm$0.02 & 97.85$\pm$0.17 \\
    LGLP & 91.30$\pm$0.05 & \underline{89.41}$\pm$0.13 & \multicolumn{1}{c}{--} & 98.51$\pm$0.01 & \multicolumn{1}{c}{--} \\
    GCN & 90.25$\pm$0.53 & 71.47$\pm$1.40 & 96.33$\pm$0.80 & 99.43$\pm$0.02 & 99.82$\pm$0.05 \\
    GSAGE & 90.24$\pm$0.34 & 87.38$\pm$1.39 & \underline{96.78}$\pm$0.11 & 99.29$\pm$0.04 & 99.93$\pm$0.02 \\
    JKNet & 89.05$\pm$0.67 & 88.58$\pm$1.78 & 96.58$\pm$0.23 & 99.43$\pm$0.02 & \textbf{\underline{99.94}}$\pm$0.01 \\
    \midrule
    \multicolumn{5}{l}{Our proposed \method with different graph encoders} \\
    \method w/ GCN & 92.55$\pm$0.50 & 89.65$\pm$0.20 & 96.99$\pm$0.08 & 99.38$\pm$0.01 & 99.44$\pm$0.05 \\
    \method w/ GSAGE & 92.61$\pm$0.52 & 91.84$\pm$0.20 & 97.01$\pm$0.01 & 99.34$\pm$0.10 & 99.83$\pm$0.05 \\
    \method w/ JKNet & \textbf{93.05}$\pm$0.24 & \textbf{92.12}$\pm$0.47 & \textbf{97.53}$\pm$0.17 & 99.31$\pm$0.04 & \textbf{99.94}$\pm$0.01 \\
    \bottomrule
  \end{tabular}
  \vspace{-0.05in}
\end{table*}

\subsection{Experimental Setup}
\label{sec:exp_setup}

We conduct experiments on five benchmark datasets including citation networks (\cora, \cseer, \pubmed~\citep{yang2016revisiting}), social network (\facebook~\citep{mcauley2012learning}), and drug-drug interaction network (\ddi~\citep{wishart2018drugbank}) from the Open Graph Benchmark (OGB)~\citep{hu2020open}. 
For the first four datasets, we randomly select 10\%/20\% of the links and the same numbers of disconnected node pairs as validation/test samples. The links in the validation and test sets are masked off from the training graph. For \ddi, we used the OGB official train/validation/test splits. 
Statistics for the datasets are shown in \cref{tab:data}, with more details in Appendix.
We use K-core~\citep{bader2003automated} clusters as the default treatment variable. We evaluate \method on three commonly used GNN encoders: GCN~\citep{kipf2016semi}, GSAGE~\citep{hamilton2017inductive}, and JKNet~\citep{xu2018representation}.
We compare the link prediction performance of \method against 
Node2Vec~\citep{grover2016node2vec}, MVGRL~\citep{hassani2020contrastive}, VGAE~\citep{kipf2016variational}, SEAL~\citep{zhang2018link}, LGLP~\citep{cai2021line}, and GNNs with MLP decoder. We report averaged test performance and their standard deviation over 20 runs with different random parameter initializations. Other than the most commonly used of Area Under ROC Curve (AUC), we report Hits@20 (one of the primary metrics on OGB leaderboard) as a more challenging metric, as it expects models to rank positive edges higher than nearly all negative edges. 

Besides performance comparison on link prediction, we will answer two questions to suggest a way of choosing a treatment variable for creating counterfactual links: (Q1) Does \method sufficiently learn the observed \emph{averaged treatment effect} (ATE) derived from the counterfactual links? (Q2) What is the relationship between the estimated ATE learned in the method and the prediction performance?
If the answer to Q1 is yes, then the answer to Q2 will indicate how to choose treatment based on observed ATE.
To answer the Q1, we calculate the observed ATE ($\widehat{\text{ATE}}_{obs}$) by comparing the observed links in $\mathbf{A}$ and created counterfactual links $\mathbf{A}^{CF}$ that have opposite treatments.
And we calculate the estimated ATE ($\widehat{\text{ATE}}_{est}$) by comparing the predicted links in $\widehat{\mathbf{A}}$ and predicted counterfactual links $\widehat{\mathbf{A}}^{CF}$.
Formally, $\widehat{\text{ATE}}_{obs}$ and $\widehat{\text{ATE}}_{est}$ are defined as
\begin{align}
    \label{eq:ate_obs}
    \widehat{\text{ATE}}_{obs} = &\frac{1}{N^2} \sum_{i=1}^{N} \sum_{j=1}^{N} {\{} \mathbf{T} \odot (\mathbf{A} - \mathbf{A}^{CF}) \\
    &+ (\mathbf{1}_{N \times N} - \mathbf{T}) \odot (\mathbf{A}^{CF} - \mathbf{A}) {\}}_{i,j}. \nonumber \\
    \label{eq:ate_est}
    \widehat{\text{ATE}}_{est} = &\frac{1}{N^2} \sum_{i=1}^{N} \sum_{j=1}^{N} {\{} \mathbf{T} \odot (\widehat{\mathbf{A}} - \widehat{\mathbf{A}}^{CF}) \\
    &+ (\mathbf{1}_{N \times N} - \mathbf{T}) \odot (\widehat{\mathbf{A}}^{CF} - \widehat{\mathbf{A}}) {\}}_{i,j}. \nonumber
\end{align}
The treatment variables we will investigate are generally graph clustering or community detection methods, such as K-core~\citep{bader2003automated}, stochastic block model (SBM)~\citep{karrer2011stochastic}, spectral clustering (SpecC)~\citep{ng2001spectral}, propagation clustering (PropC)~\citep{raghavan2007near}, Louvain~\citep{blondel2008fast}, common neighbors (CommN), Katz index, and hierarchical clustering (Ward)~\citep{ward1963hierarchical}. We use JKNet~\citep{xu2018representation} as default graph encoder.

Implementation details and supplementary experimental results (e.g., sensitivity on $\gamma$, ablation study on $\mathcal{L}_{CF}$ and $\mathcal{L}_{disc}$) can be found in Appendix. Source code is available in supplementary material.

\subsection{Experimental Results}

\paragraph{Link Prediction}
\cref{tab:result_hits,tab:result_auc} show the link prediction performance of Hits@20 and AUC by all methods. LGLP on \pubmed and \ddi are missing due to the out of memory error when running the official code package from the authors. We observe that our \method on different graph encoders achieve similar or better performances compared with baselines. The only exception is the AUC on \facebook where most methods have close-to-perfect AUC.
As AUC is a relatively easier metric comparing with Hits@20, most methods achieved good performance on AUC. 
We observe that \method with JKNet almost consistently achieves the best performance and outperforms baselines significantly on Hits@20. Specifically, comparing with the best baseline, \method improves relatively by 16.4\% and 0.8\% on Hits@20 and AUC, respectively.
Comparing with the best performing baselines, which are also GNN-based, \method benefits from learning with both observed link existence ($\mathbf{A}$) and our defined counterfactual links ($\mathbf{A}^{CF}$).

\vspace{-0.04in}
\paragraph{ATE with Different Treatments}
\cref{tab:ate_cora,tab:ate_cseer} show the link prediction performance, $\widehat{\text{ATE}}_{obs}$, and $\widehat{\text{ATE}}_{est}$ of \method (with JKNet) when using different treatments. The treatments in \cref{tab:ate_cora,tab:ate_cseer} are sorted by the Hits@20 performance. Bigger ATE indicates stronger causal relationship between the treatment and outcome, and vice versa. We observe: 
(1) the rankings of $\widehat{\text{ATE}}_{est}$ and $\widehat{\text{ATE}}_{obs}$ are positively correlated with Kendell's ranking coefficient~\citep{abdi2007kendall} of 0.67 and 0.57 for \cora and \cseer, respectively. Hence, \method was sufficiently trained to learn the causal relationship between graph structure information and link existence;
(2) $\widehat{\text{ATE}}_{obs}$ and $\widehat{\text{ATE}}_{est}$ are both negatively correlated with the link prediction performance, showing that we can pick a proper treatment prior to training a model with \method. Using the treatment that has the weakest causal relationship with link existence is likely to train the model to capture more essential factors on the outcome, in a way similar to denoising the unrelated information from the representations.
While methods that learn from only observed
data may assume strongly positive correlation for this treatment, the counterfactual data are more useful to complement the partial observations for learning better representations.

\begin{table}[t]
\small
\centering
\caption{Results of \method with different treatments on \cora. (sorted by Hits@20)}
\label{tab:ate_cora}
  \begin{tabular}{lcccc}
    \toprule
    & Hits@20 & $\widehat{\text{ATE}}_{obs}$ & $\widehat{\text{ATE}}_{est}$ \\
    \midrule
    K-core & 65.6$\pm$1.1 & 0.002 & 0.013$\pm$0.003 \\
    SBM & 64.2$\pm$1.1 & 0.006 & 0.023$\pm$0.015 \\
    CommN & 62.3$\pm$1.6 & 0.007 & 0.053$\pm$0.021 \\
    PropC & 61.7$\pm$1.4 & 0.037 & 0.059$\pm$0.065 \\
    Ward & 61.2$\pm$2.3 & 0.001 & 0.033$\pm$0.012 \\
    SpecC & 59.3$\pm$2.8 & 0.002 & 0.033$\pm$0.011 \\
    Louvain & 57.6$\pm$1.8 & 0.025 & 0.138$\pm$0.091 \\
    Katz & 56.6$\pm$3.4 & 0.740 & 0.802$\pm$0.041 \\
    \bottomrule
  \end{tabular}
\vspace{-0.05in}
\end{table}
 
\begin{table}[t]
\small
\centering
\caption{Results of \method with different treatments on \cseer. (sorted by Hits@20)}
\label{tab:ate_cseer}
  \begin{tabular}{lcccc}
    \toprule
    & Hits@20 & $\widehat{\text{ATE}}_{obs}$ & $\widehat{\text{ATE}}_{est}$ \\
    \midrule
    SBM & 71.6 $\pm$1.9 & 0.004 & 0.005 $\pm$0.001 \\
    K-core & 68.1$\pm$1.5 & 0.002 & 0.010$\pm$0.002 \\
    Ward & 67.0$\pm$1.7 & 0.003 & 0.037$\pm$0.009 \\
    PropC & 64.6$\pm$3.6 & 0.141 & 0.232$\pm$0.113 \\
    Louvain & 63.3$\pm$2.5 & 0.126 & 0.151$\pm$0.078 \\
    SpecC & 59.9$\pm$1.3 & 0.009 & 0.166$\pm$0.034 \\
    Katz & 57.3$\pm$0.5 & 0.245 & 0.224$\pm$0.037 \\
    CommN & 56.8$\pm$4.9 & 0.678 & 0.195$\pm$0.034 \\
    \bottomrule
  \end{tabular}
\vspace{-0.05in}
\end{table}

%% file: body_icml/6related.tex
\paragraph{Link Prediction} With its wide applications, link prediction has drawn attention from many research communities including statistical machine learning and data mining. Stochastic generative methods based on stochastic block models (SBM) are developed to generate links~\citep{mehta2019stochastic}. In data mining, matrix factorization~\citep{menon2011link}, heuristic methods~\citep{philip2010link,martinez2016survey}, and graph embedding methods~\citep{cui2018survey} have been applied to predict links in the graph. Heuristic methods compute the similarity score of nodes based on their neighborhoods. These methods can be generally categorized into first-order, second-order, and high-order heuristics based on the maximum distance of the neighbors. Graph embedding methods learn latent node features via embedding lookup and use them for link prediction~\citep{perozzi2014deepwalk,tang2015line,grover2016node2vec,wang2016structural}.

In the past few years, GNNs have shown promising results on various graph-based tasks with their ability of learning from features and custom aggregations on structures~\citep{kipf2016semi,hamilton2017inductive,ma2021unified,jiang2022federated}. With node pair representations and an attached MLP or inner-product decoder, GNNs can be used for link prediction~\citep{davidson2018hyperspherical,yang2018binarized,zhang2020revisiting,yun2021neo,zhu2021neural,wang2021modeling,wang2021dynamic}.
For example, VGAE used GCN to learn node representations and reconstruct the graph structure~\citep{kipf2016variational}. SEAL extracted a local subgraph around each target node pair and then learned local subgraph representation for link prediction~\citep{zhang2018link}. Following the scheme of SEAL, \citet{cai2020multi} proposed to improve local subgraph representation learning by multi-scale graph representation. And LGLP proposed to invert the local subgraphs to line graphs~\citep{cai2021line}.
However, little work has studied to use causal inference for improving link prediction.

\paragraph{Causal Inference}
Causal inference methods usually re-weighted samples based on propensity score~\citep{rosenbaum1983central,austin2011introduction} to remove confounding bias from binary treatments. Recently, several works studied about learning treatment invariant representation to predict the counterfactual outcomes~\citep{shalit2017estimating,li2017matching,yao2018representation,yoon2018ganite,hassanpour2019counterfactual,hassanpour2019learning,bica2020estimating}.
Few recent works combined causal inference with graph learning~\citep{sherman2020intervening,bevilacqua2021size,lin2021generative,feng2021should}. For example, \citet{sherman2020intervening} proposed network intervention to study the effect of link creation on network structure changes.

As a mean of learning the causality between treatment and outcome, counterfactual prediction has been used for a variety of applications such as recommender systems~\citep{wang2020information,xu2020adversarial}, health care~\citep{alaa2017bayesian,pawlowski2020deep}, 
and decision making~\citep{kusner2017counterfactual,pitis2020counterfactual}.
To infer the causal relationships, previous work usually estimated the ITE via function fitting models~\citep{kuang2017treatment,wager2018estimation,kuang2019treatment,assaad2021counterfactual}.

\paragraph{Graph Data Augmentation}
Graph data augmentation (GDA) methods generate perturbed or modified graph data~\cite{zhao2021data,zhao2021action} to improve the generalizability of graph machine learning models. Two comprehensive surveys of graph data augmentation are given by \citet{zhao2022graph} and \citet{ding2022data}. So far, most GDA methods have been focusing on node-level tasks~\citep{park2021metropolis} and graph-level tasks~\citep{liu2022graph,luo2022automated}.
Due to the non-Euclidean structure of graphs, most GDA work focused on modifying the graph structure. E.g., edge dropping methods~\citep{rong2019dropedge,zheng2020robust,luo2021learning} drop edges during training to reduce overfitting. 
\citet{zhao2021data} used link predictor to manipulate the graph structure and improve the graph's homophily. 
Recently, several works also combined GDA with self-supervised learning objectives such as contrastive learning~\cite{you2020graph,you2021graph,zhu2021graph} and consistency loss~\cite{wang2020nodeaug,feng2020graph}. 
Nevertheless, GDA for link prediction has been under-explored.

%% file: body_icml/7conclusion.tex
In this work, we presented the novel concept of counterfactual link and a novel graph learning method for link prediction (\method). The counterfactual links answered the counterfactual questions on the link existence and were used as augmented training data, with which \method accurately predicted missing links by exploring the causal relationship between global graph structure and link existence. Extensive experiments demonstrated that \method achieved the state-of-the-art performance on benchmark datasets. 
This work sheds insights that a good use of causal models (even basic ones) can greatly improve the performance of (graph) machine learning tasks such as link prediction. We note that the use of more sophistically designed causal models may lead to larger improvements for machine learning tasks, which can be a valuable future direction for the research community. Other than cluster-based global graph structure as treatment, other choices (with both empirical and theoretical analyses) are also worthy of exploration.

%% file: body_icml/8appendix.tex
\section{Additional Dataset Details}
In this section, we provide some additional dataset details. All the datasets used in this work are publicly available. 

\paragraph{Citation Networks}
\cora, \cseer, and \pubmed are citation networks that were first used by \citet{yang2016revisiting} and then commonly used as benchmarks in GNN-related literature~\citep{kipf2016semi,velivckovic2017graph}. In these citation networks, the nodes are published papers and features are bag-of-word vectors extracted from the corresponding paper. Links represent the citation relation between papers. We loaded the datasets with the \verb+DGL+\footnote{\url{https://github.com/dmlc/dgl}} package.

\paragraph{Social Network} The \facebook dataset\footnote{\url{https://snap.stanford.edu/data/ego-Facebook.html}} is a social network constructed from friends lists from Facebook~\citep{mcauley2012learning}. The nodes are Facebook users and links indicate the friendship relation on Facebook. The node features were constructed from the user profiles and anonymized by \citet{mcauley2012learning}. 

\paragraph{Drug-Drug Interaction Network} The \ddi dataset was constructed from a public Drug database~\citep{wishart2018drugbank} and provided by the Open Graph Benchmark (OGB)~\citep{hu2020open}. Each node in this graph represents an FDA-approved or experimental drug and edges represent the existence of unexpected effect when the two drugs are taken together. This dataset does not contain any node features, and it can be downloaded with the dataloader\footnote{\url{https://ogb.stanford.edu/docs/linkprop/\#data-loader}} provided by OGB.

\section{Details on Implementation and Hyperparameters}
All the experiments in this work were conducted on a Linux server with Intel Xeon Gold 6130 Processor (16 Cores @2.1Ghz), 96 GB of RAM, and 4 RTX 2080Ti cards (11 GB of RAM each). Our method are implemented with \verb+Python 3.8.5+ with \verb+PyTorch+. Source code is publicly available at \url{https://github.com/DM2-ND/CFLP}.

\paragraph{Baseline Methods}
For baseline methods, we use official code packages from the authors for MVGRL\footnote{\url{https://github.com/kavehhassani/mvgrl}}~\citep{hassani2020contrastive}, SEAL\footnote{\url{https://github.com/facebookresearch/SEAL_OGB}}~\citep{zhang2018link}, and LGLP\footnote{\url{https://github.com/LeiCaiwsu/LGLP}}~\citep{cai2021line}. We use a public implementation for VGAE\footnote{\url{https://github.com/DaehanKim/vgae_pytorch}}~\citep{kipf2016variational} and OGB implementations\footnote{\url{https://github.com/snap-stanford/ogb/tree/master/examples/linkproppred/ddi}} for Node2Vec and baseline GNNs. For fair comparison, we set the size of node/link representations to be 256 of all methods.

\paragraph{\method} 
We use the Adam optimizer with a simple cyclical learning rate scheduler~\citep{smith2017cyclical}, in which the learning rate waves cyclically between the given learning rate ($lr$) and 1e-4 in every 70 epochs (50 warmup steps and 20 annealing steps). We implement the GNN encoders with \verb+torch_geometric+\footnote{\url{https://pytorch-geometric.readthedocs.io/en/latest/}}~\citep{fey2019fast}. Same with the baselines, we set the size of all hidden layers and node/link representations of \method as 256. The graph encoders all have three layers and JKNet has mean pooling for the final aggregation layer. The decoder is a 3-layer MLP with a hidden layer of size 64 and ELU as the nonlinearity. As the Euclidean distance used in \cref{eq:nearest2} has a range of $[0, \infty)$, the value of $\gamma$ depends on the distribution of all-pair node embedding distances, which varies for different datasets. Therefore, we set the value of $\gamma$ as the $\gamma_{pct}$-percentile of all-pair node embedding distances. Commands for reproducing the experiments are included in \verb+README.md+.

\paragraph{Hyperparameter Searching Space} We manually tune the following hyperparameters over range: $lr \in \{0.005, 0.01, 0.05, 0.1, 0.2\}$, $\alpha \in \{0.001, 0.01, 0.1, 1, 2\}$, $\beta \in \{0.001, 0.01, 0.1, 1, 2\}$, $\gamma_{pct} \in \{10, 20, 30\}$.

\paragraph{Treatments}
For the graph clustering or community detection methods we used as treatments, we use the implementation from \verb+scikit-network+\footnote{\url{https://scikit-network.readthedocs.io/}} for Louvain~\citep{blondel2008fast}, SpecC~\citep{ng2001spectral}, PropC~\citep{raghavan2007near}, and Ward~\citep{ward1963hierarchical}.
We used implementation of K-core~\citep{bader2003automated} from \verb+networkx+.\footnote{\url{https://networkx.org/documentation/}}
We used SBM~\citep{karrer2011stochastic} from a public implementation by \citet{funke2019stochastic}.\footnote{\url{https://github.com/funket/pysbm}}
For CommN and Katz, we set $T_{i, j} = 1$ if the number of common neighbors or Katz index between $v_i$ and $v_j$ are greater or equal to 2 or 2 times the average of all Katz index values, respectively.
For SpecC, we set the number of clusters as 16.
For SBM, we set the number of communities as 16. 
These settings are fixed for all datasets.

\begin{table}[t]
\small
  \caption{Link prediction performances measured by Hits@50. Best performance and best baseline performance are marked with bold and underline, respectively.}
  \label{tab:result_hits50}
  \centering
  \begin{tabular}{lccccc}
    \toprule
    & \cora & \cseer & \pubmed & \facebook & \ddi \\
    \midrule
    Node2Vec & 63.64$\pm$0.76 & 54.57$\pm$1.40 & 50.73$\pm$1.10 & 43.91$\pm$1.03 & 24.34$\pm$1.67 \\
    MVGRL & 29.97$\pm$3.06 & 26.48$\pm$0.98 & 16.96$\pm$0.56 & 17.06$\pm$0.19 & 12.03$\pm$0.11 \\
    VGAE & 60.36$\pm$2.71 & 54.68$\pm$3.15 & 41.98$\pm$0.31 & 51.36$\pm$0.93 & 23.00$\pm$1.66 \\
    SEAL & 51.68$\pm$2.85 & 54.55$\pm$1.77 & 42.85$\pm$2.03 & 57.20$\pm$1.85 & 40.85$\pm$2.97 \\
    LGLP & \underline{71.43}$\pm$0.75 & \underline{69.98}$\pm$0.16 & -- & 56.22$\pm$0.49 & -- \\
    GCN & 64.93$\pm$1.62 & 63.38$\pm$1.73 & 39.20$\pm$6.47 & \underline{69.90}$\pm$0.65 & 73.70$\pm$3.99 \\
    GSAGE & 63.18$\pm$3.39 & 61.71$\pm$2.43 & \underline{54.81}$\pm$2.67 & 62.53$\pm$4.24 & 86.83$\pm$3.85 \\
    JKNet & 62.64$\pm$1.40 & 62.26$\pm$2.10 & 45.16$\pm$3.18 & 68.81$\pm$1.76 & \underline{91.48}$\pm$2.41 \\
    \midrule
    \multicolumn{6}{l}{Our proposed \method with different graph encoders} \\
    \method w/ GCN & 72.61$\pm$0.92 & 69.85$\pm$1.11 & 55.00$\pm$1.95 & 70.47$\pm$0.77 & 62.47$\pm$1.53 \\
    \method w/ GSAGE & 73.25$\pm$0.94 & 64.75$\pm$2.27 & 58.16$\pm$1.40 & 63.89$\pm$2.08 & 83.32$\pm$3.61 \\
    \method w/ JKNet & \textbf{75.49}$\pm$1.54 & \textbf{77.01}$\pm$1.92 & \textbf{62.80}$\pm$0.79 & \textbf{71.41}$\pm$0.61 & \textbf{93.07}$\pm$1.14 \\
    \bottomrule
  \end{tabular}
\end{table}

\begin{table}[t]
\small
  \caption{Link prediction performances measured by Average Precision (AP). Best performance and best baseline performance are marked with bold and underline, respectively.}
  \label{tab:result_ap}
  \centering
  \begin{tabular}{lccccc}
    \toprule
    & \cora & \cseer & \pubmed & \facebook & \ddi \\
    \midrule
    Node2Vec & 88.53$\pm$0.42 & 84.42$\pm$0.48 & 87.15$\pm$0.12 & 99.07$\pm$0.02 & 98.39$\pm$0.04 \\
    MVGRL & 76.47$\pm$3.07 & 67.40$\pm$0.52 & 82.00$\pm$0.97 & 82.37$\pm$0.35 & 81.12$\pm$1.77 \\
    VGAE & 89.89$\pm$0.50 & 86.97$\pm$0.78 & 95.97$\pm$0.16 & 98.60$\pm$0.04 & 95.28$\pm$0.11 \\
    SEAL & 89.08$\pm$0.57 & 88.55$\pm$0.32 & 96.33$\pm$0.28 & \textbf{\underline{99.51}}$\pm$0.03 & 98.39$\pm$0.21 \\
    LGLP & \underline{93.05}$\pm$0.03 & \underline{91.62}$\pm$0.09 & -- & 98.62$\pm$0.01 & -- \\
    GCN & 91.42$\pm$0.45 & 90.87$\pm$0.52 & 96.19$\pm$0.88 & 99.42$\pm$0.02 & 99.86$\pm$0.03 \\
    GSAGE & 91.52$\pm$0.46 & 89.43$\pm$1.15 & \underline{96.93}$\pm$0.11 & 99.27$\pm$0.06 & 99.93$\pm$0.01 \\
    JKNet & 90.50$\pm$0.22 & 90.42$\pm$1.34 & 96.56$\pm$0.31 & 99.41$\pm$0.02 & \underline{99.95}$\pm$0.01 \\
    \midrule
    \multicolumn{6}{l}{Our proposed \method with different graph encoders} \\
    \method w/ GCN & 93.77$\pm$0.49 & 91.84$\pm$0.20 & 97.16$\pm$0.08 & 99.40$\pm$0.01 & 99.60$\pm$0.03 \\
    \method w/ GSAGE & 93.55$\pm$0.49 & 90.80$\pm$0.87 & 97.10$\pm$0.08 & 99.29$\pm$0.06 & 99.88$\pm$0.04 \\
    \method w/ JKNet & \textbf{94.24}$\pm$0.28 & \textbf{93.92}$\pm$0.41 & \textbf{97.69}$\pm$0.13 & 99.35$\pm$0.02 & \textbf{99.96}$\pm$0.01 \\
    \bottomrule
  \end{tabular}
\end{table}

\begin{table}[t]
\small
  \caption{Link prediction performance of \method (w/ JKNet) on \cora and \cseer when removing $\mathcal{L}_{CF}$ or $\mathcal{L}_{disc}$ or both versus normal setting.}
  \label{tab:result_ablation}
  \centering
  \begin{tabular}{lcc|cc}
    \toprule
    & \multicolumn{2}{c}{\cora} & \multicolumn{2}{c}{\cseer} \\
    & Hits@20 & AUC & Hits@20 & AUC \\
    \midrule
    \method ($\alpha=0$) &  58.58$\pm$0.23 & 89.16$\pm$0.93 & 65.49$\pm$2.18 & 91.01$\pm$0.64 \\
    \method ($\beta=0$) & 62.27$\pm$0.84 & 92.96$\pm$0.34 & 66.92$\pm$1.84 & 91.98$\pm$0.17 \\
    \method ($\alpha=\beta=0$) & 58.52$\pm$0.83 & 88.79$\pm$0.28 & 64.69$\pm$3.25 & 90.61$\pm$0.64 \\
    \method & \textbf{65.57}$\pm$1.05 & \textbf{93.05}$\pm$0.24 & \textbf{68.09}$\pm$1.49 & \textbf{92.12}$\pm$0.47 \\
    \bottomrule
  \end{tabular}
  \vspace{-0.1in}
\end{table}

\begin{table}[t]
\small
  \caption{Link prediction performance of \method (w/ JKNet) on \cora and \cseer with node embeddings ($\tilde{\mathbf{X}}$) learned from different methods.}
  \label{tab:emb_ablation}
  \centering
  \begin{tabular}{lcc|cc|cc}
    \toprule
    & \multicolumn{2}{c}{\cora} & \multicolumn{2}{c}{\cseer} & \multicolumn{2}{c}{\ddi} \\
    & Hits@20 & AUC & Hits@20 & AUC & Hits@20 & AUC \\
    \midrule
    (MVGRL) & 65.57$\pm$1.05 & 93.05$\pm$0.24 & 68.09$\pm$1.49 & 92.12$\pm$0.47 & 86.08$\pm$1.98 & 99.94$\pm$0.01 \\
    (GRACE) & 62.54$\pm$1.41 & 92.28$\pm$0.69 & 68.68$\pm$1.75 & 93.80$\pm$0.36 & 82.30$\pm$3.32 & 99.93$\pm$0.01 \\
    (DGI) & 61.04$\pm$1.52 & 92.99$\pm$0.49 & 72.17$\pm$1.08 & 93.34$\pm$0.51 & 85.61$\pm$1.66 & 99.94$\pm$0.01 \\
    \bottomrule
  \end{tabular}
\end{table}


\section{Additional Experimental Results and Discussions}

\paragraph{Link Prediction}
\cref{tab:result_hits50,tab:result_ap} show the link prediction performance of Hits@50 and Average Precision (AP) by all methods. LGLP on \pubmed and \ddi are missing due to the out of memory error when running the code package from the authors. Similar to the results in \cref{tab:result_hits,tab:result_auc}, we observe that our \method on different graph encoders achieve similar or better performances compared with baselines, with the only exception of AP on \facebook where most methods have close-to-perfect AP.
From \cref{tab:result_hits,tab:result_auc,tab:result_hits50,tab:result_ap}, we observe that \method achieves improvement over all GNN architectures (averaged across datasets). Specifically, CFLP improves 25.6\% (GCN), 12.0\% (GSAGE), and 36.3\% (JKNet) on Hits@20, 9.6\% (GCN), 5.0\% (GSAGE), and 17.8\% (JKNet) on Hits@50, 5.6\% (GCN), 1.6\% (GSAGE), and 1.9\% (JKNet) on AUC, and 0.8\% (GCN), 0.8\% (GSAGE), and 1.8\% (JKNet) on AP.
We note that \method with JKNet almost consistently achieves the best performance and outperforms baselines significantly on Hits@50. Specifically, compared with the best baseline, \method improves relatively by 6.8\% and 0.9\% on Hits@50 and AP, respectively.

\paragraph{Ablation Study on Losses}
For the ablative studies of $\mathcal{L}_{CF}$ (\cref{eq:eplosscf}) and $\mathcal{L}_{disc}$ (\cref{eq:discloss}), we show their effect by removing them from the integrated loss function (\cref{eq:loss}). 
\cref{tab:result_ablation} shows the results of \method on \cora and \cseer under different settings ($\alpha=0$, $\beta=0$, $\alpha = \beta = 0$, and original setting).
We observe that \method in the original setting achieves the best performance. The performance drops significantly when having $\alpha=0$, i.e., not using any counterfactual data during training. We note that having $\beta=0$, i.e., not using the discrepancy loss, also lowers the performance. Therefore, both $\mathcal{L}_{CF}$ and $\mathcal{L}_{disc}$ are essential for improving the link prediction performance.

\paragraph{Ablation Study on Node Embedding $\tilde{\mathbf{X}}$}
As the node embedding $\tilde{\mathbf{X}}$ is used in the early step of \method for finding the counterfactual links, the quality of $\tilde{\mathbf{X}}$ may affect the later learning process. Therefore, we also evaluate \method with different state-of-the-art unsupervised graph representation learning methods: MVGRL~\citep{hassani2020contrastive}, DGI~\citep{velickovic2019deep}, and GRACE~\citep{zhu2020deep}. \cref{tab:emb_ablation} shows the link prediction performance of \method (w/ JKNet) on \cora and \cseer with different node embeddings. We observe that the choice of the method for learning $\tilde{\mathbf{X}}$ does have an impact on the later learning process as well as the link prediction performance. Nevertheless, \cref{tab:emb_ablation} shows \method's advantage can be consistently observed with different choices of methods for learning $\tilde{\mathbf{X}}$, as \method with $\tilde{\mathbf{X}}$ learned from all three methods showed promising link prediction performance.

\begin{figure}[t]
    \centering
    \subfigure[AUC performance.]
    {\includegraphics[width=0.4\linewidth]{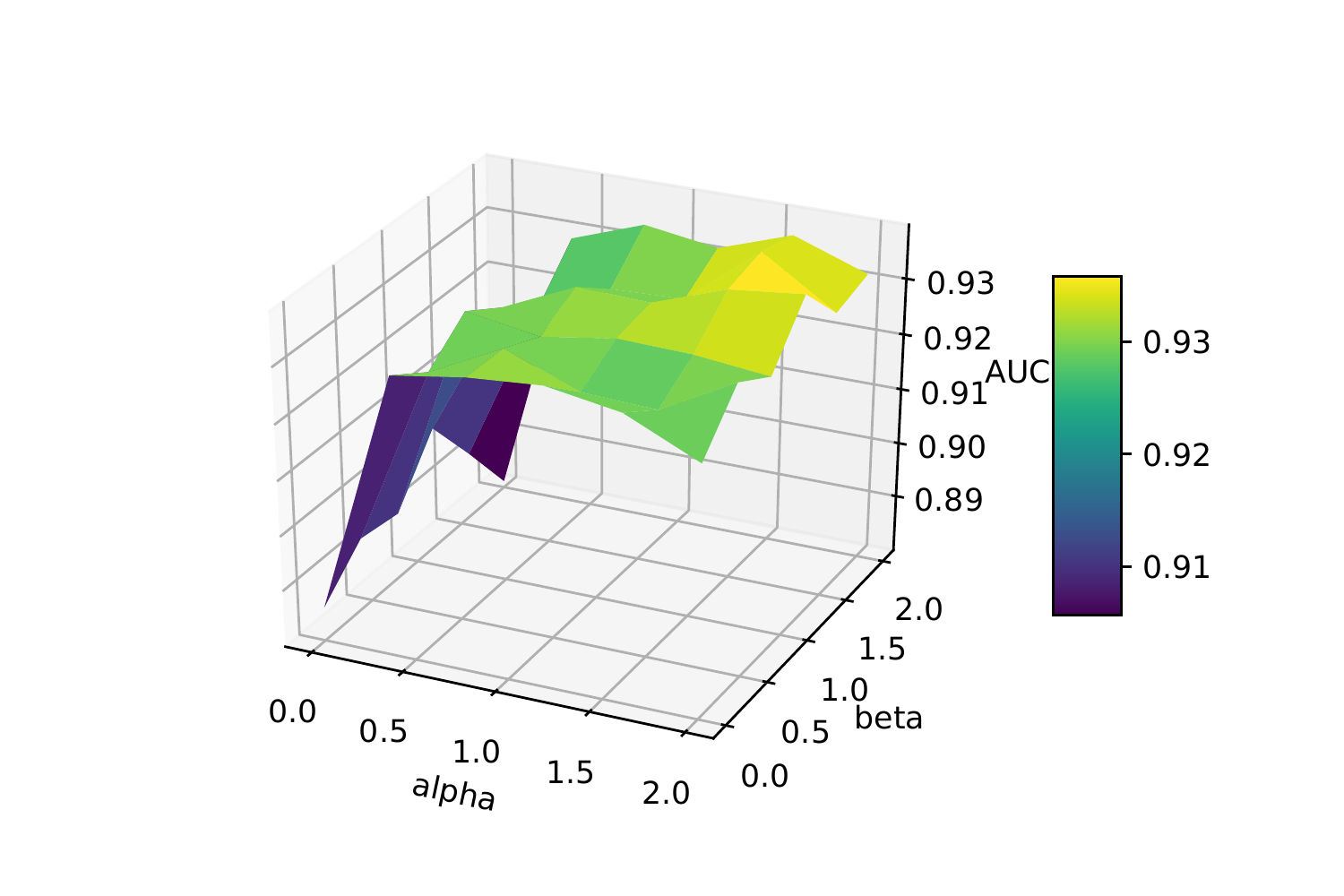}}
    \quad
    \subfigure[Hits@20 performance.]
    {\includegraphics[width=0.4\linewidth]{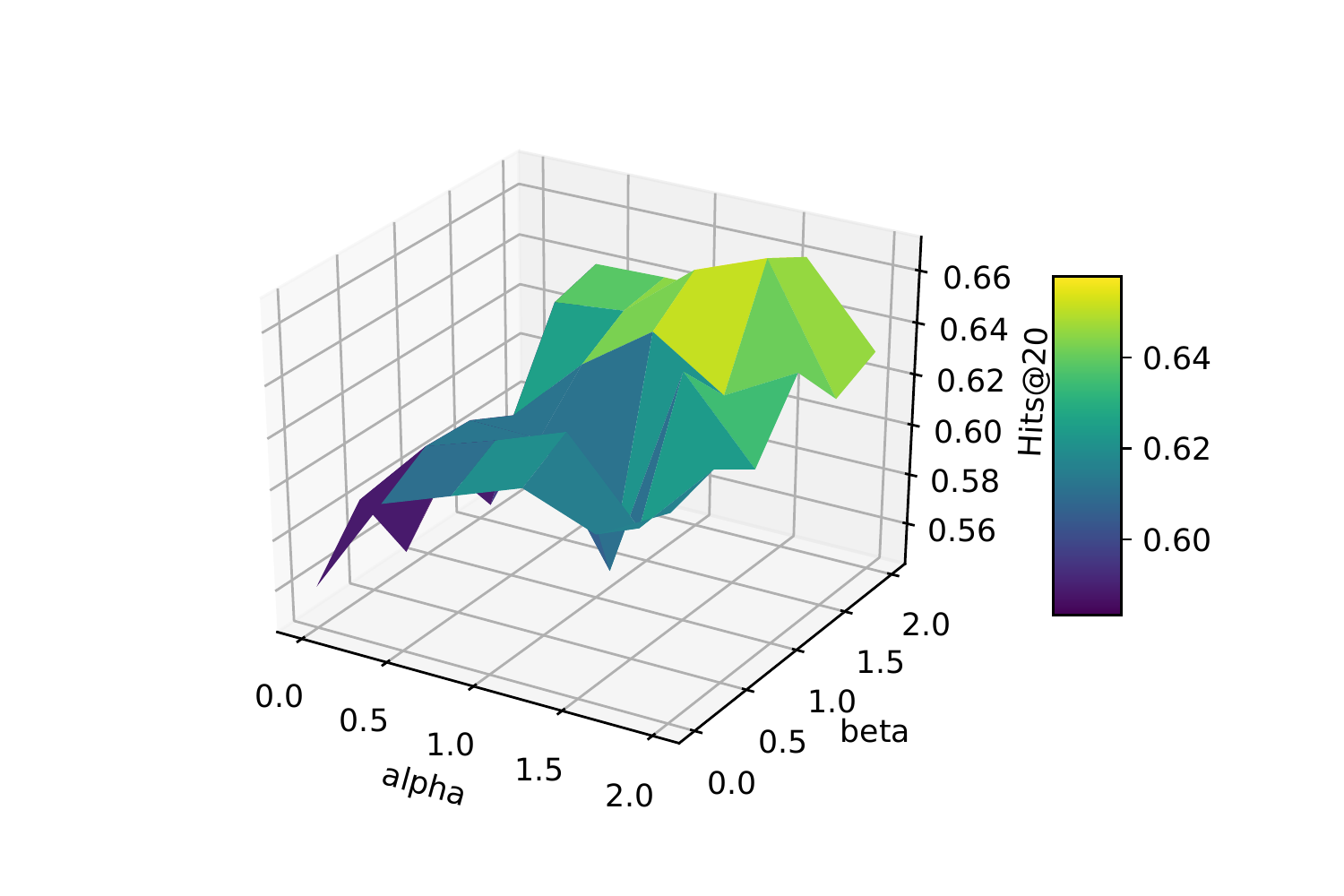}}
    \caption{Performance of \method on \cora w.r.t different combinations of $\alpha$ and $\beta$.}
    \label{fig:alphabeta}
\end{figure}

\paragraph{Sensitivity Analysis of $\alpha$ and $\beta$}
\cref{fig:alphabeta} shows the AUC performance of \method on \cora with different combinations of $\alpha$ and $\beta$. We observe that the performance is the poorest when $\alpha=\beta=0$ and gradually improves and gets stable as $\alpha$ and $\beta$ increase, showing that \method is generally robust to the hyperparameters $\alpha$ and $\beta$, and the optimal values are easy to locate.

\begin{figure}[t]
    \centering
    \subfigure[Performances of \method on \cora when using K-core as treatment.]
    {\includegraphics[width=0.4\linewidth]{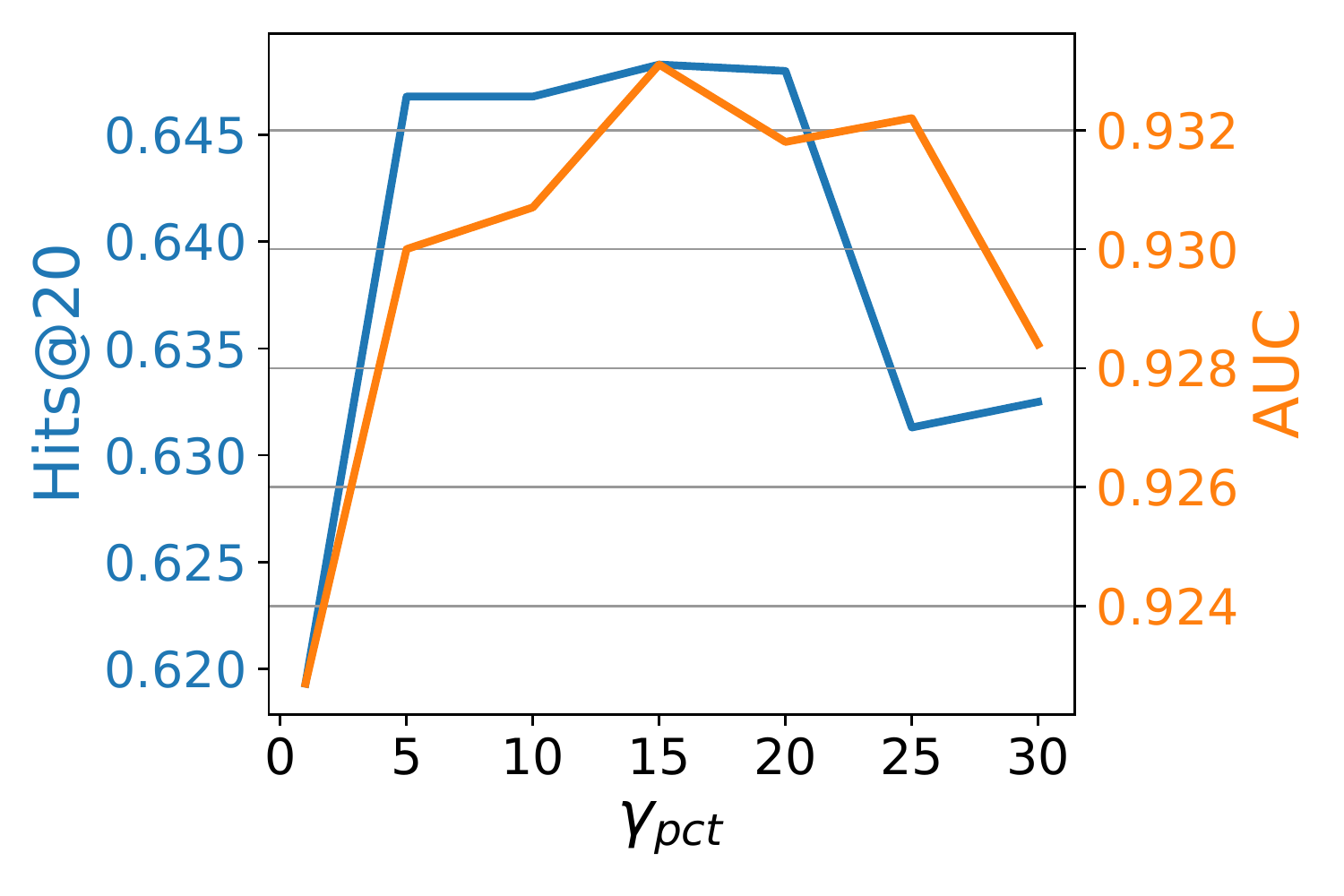}\label{fig:gamma_cora_kcore}}
    \qquad
    \subfigure[Performances of \method on \cora when using SBM as treatment.]
    {\includegraphics[width=0.4\linewidth]{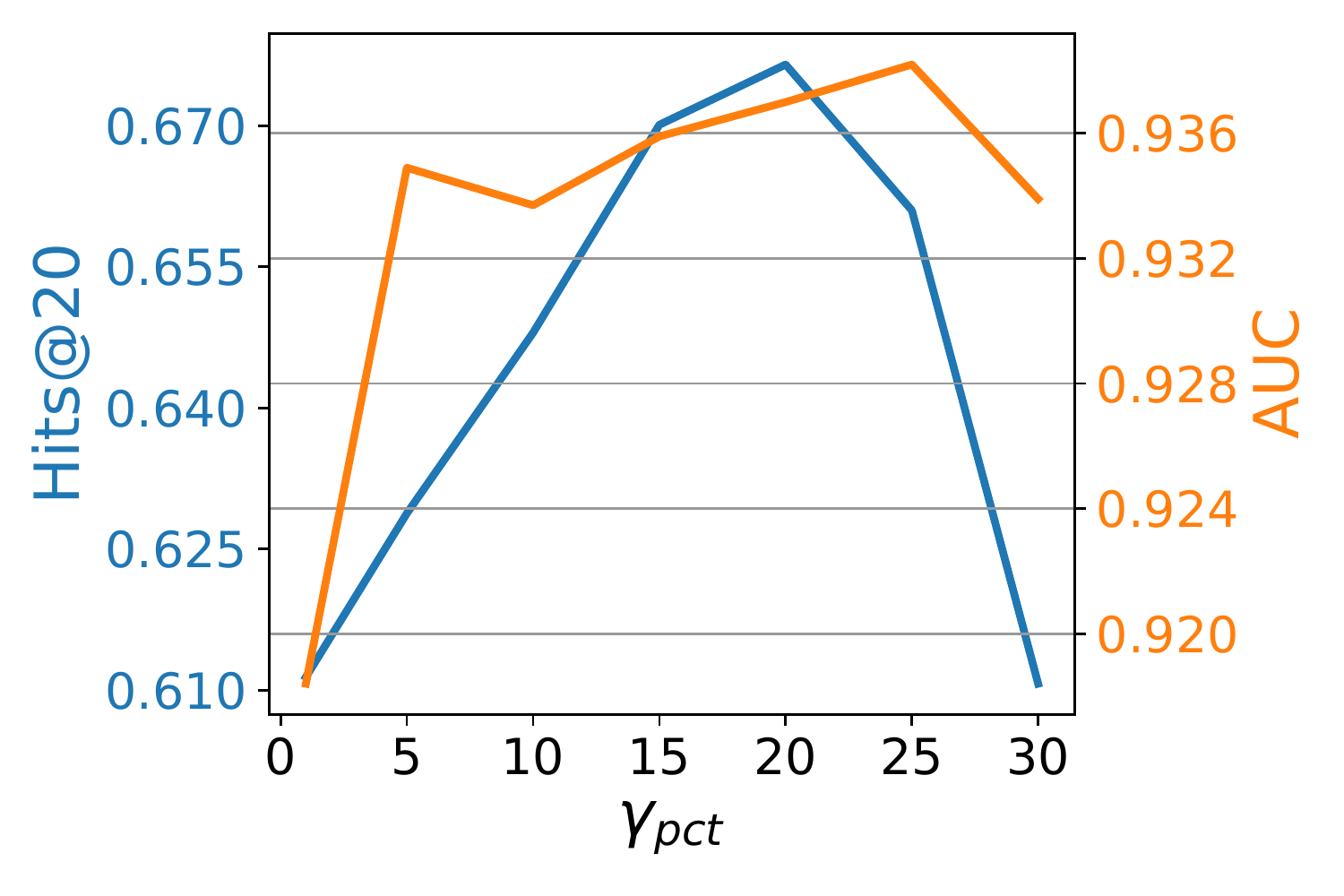}\label{fig:gamma_cora_sbm}}
    
    \subfigure[Performances of \method on \cseer when using K-core as treatment.]
    {\includegraphics[width=0.4\linewidth]{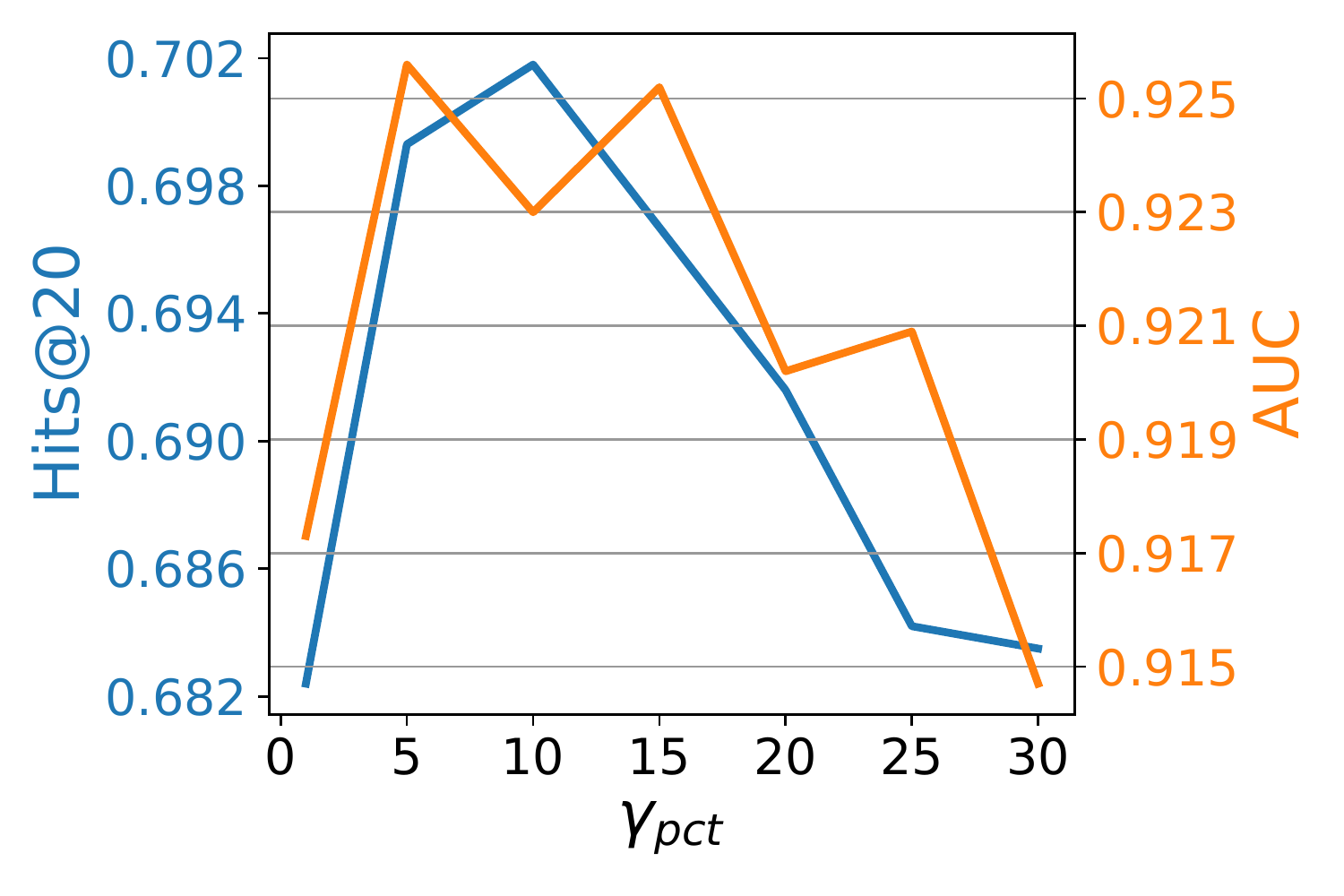}\label{fig:gamma_cseer_kcore}}
    \qquad
    \subfigure[Performances of \method on \cseer when using SBM as treatment.]
    {\includegraphics[width=0.4\linewidth]{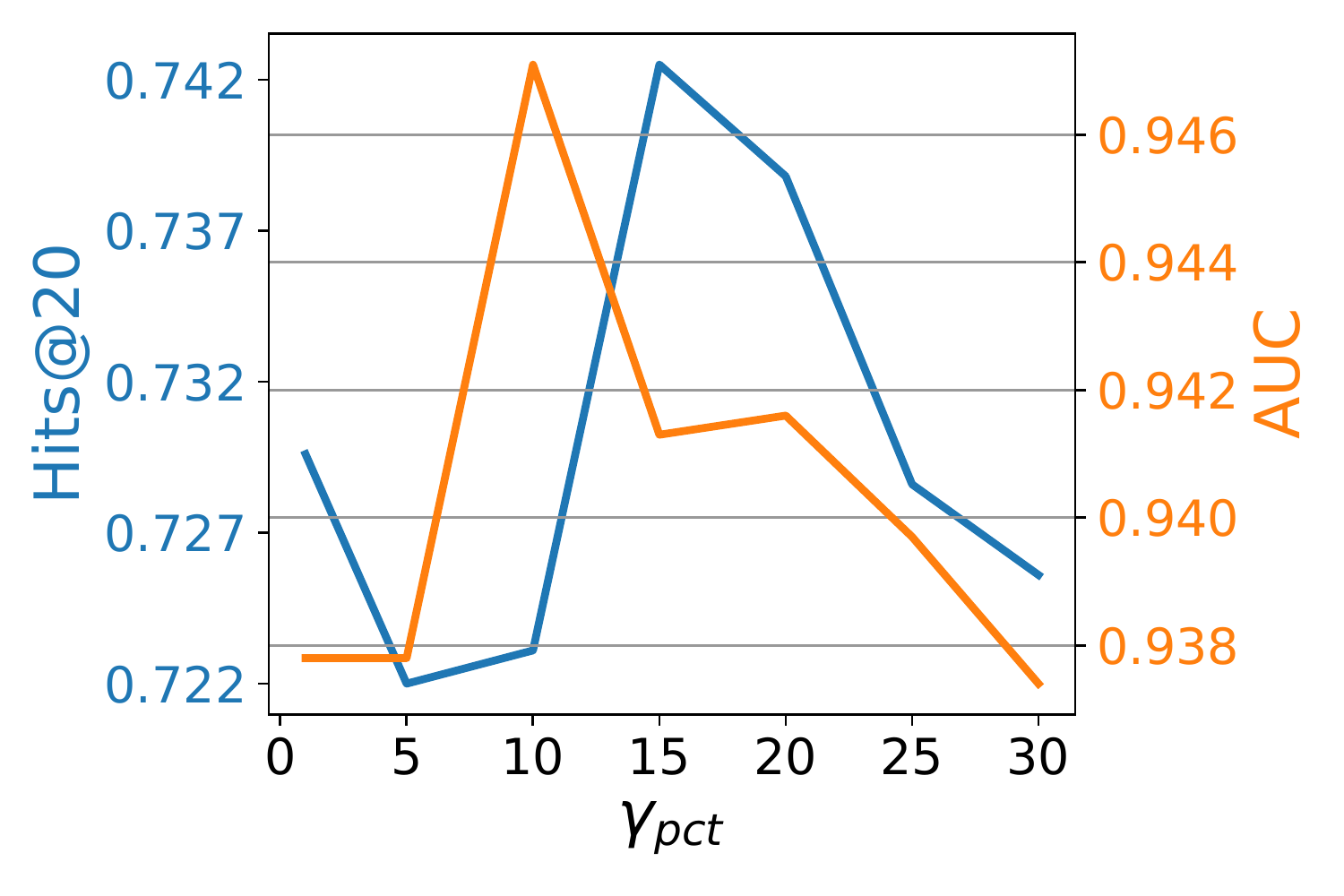}\label{fig:gamma_cseer_sbm}}
    \caption{Hits@20 and AUC performances of \method (w/ JKNet) on \cora and \cseer with different treatments w.r.t. different $\gamma_{pct}$ value.}
    \label{fig:gamma}
\end{figure}

\paragraph{Sensitivity Analysis of $\gamma$}
\cref{fig:gamma} shows the Hits@20 and AUC performance on link prediction of \method (with JKNet) on \cora and \cseer with different treatments and $\gamma_{pct}$. We observe that the performance is generally good when $10 \le \gamma_{pct} \le 20$ and gradually get worse when the value of $\gamma_{pct}$ is too small or too large, showing that \method is robust to $\gamma$ and the optimal $\gamma$ is easy to find.

